RESEARCH ARTICLE

# HukukBERT: Domain-Specific Language Model for Turkish Law

**Mehmet Utku Öztürk[1,*], Tansu Türkoğlu[2] and Buse Buz-Yalug[3]**

[1] *Kalitte Inc., Türkiye*

[2] *Aibrite Inc., Türkiye*

[3] *University of Eastern Finland, Finland*

**Abstract:** Natural language processing (NLP) advances have powered a generation of LegalTech systems, but Turkish law remains under-served by domain-specific data and models. English has legal encoders such as LEGAL-BERT; no comparable high-volume Turkish counterpart exists. We introduce HukukBERT, a Turkish legal language model trained on a 19 GB cleaned corpus using a hybrid domain-adaptive pre-training (DAPT) recipe that mixes Whole-Word Masking, Token Span Masking, Word Span Masking, and targeted Keyword Masking. We compared our 48K WordPiece tokenizer and DAPT pipeline against general-purpose and existing domain-specific Turkish models. On the Legal Cloze Test—a masked legal term prediction benchmark over Turkish court decisions—HukukBERT reaches 84.40% Top-1 accuracy and beats every baseline we tested. The Legal Cloze Test is synthetically constructed, so its passages are absent from the pre-training corpus by construction, eliminating train–test contamination. On the downstream task of structural segmentation of official Turkish court decisions, it reaches a 92.8% document pass rate. We release HukukBERT to support Turkish legal NLP work in named entity recognition, judgment prediction, and document classification.

## 1. Introduction

Transformer-based pre-trained language models [1]—most prominently BERT [2] and successors like RoBERTa [3] and DeBERTa [4]—now set the bar across most natural language processing (NLP) tasks. But applying these general-domain models to law typically falls short. Legal prose is archaic, syntactically dense, and saturated with domain phrasing that everyday corpora do not adequately reflect.

**Problem statement**. This work addresses a specific, concrete gap: Turkish law lacks both a high-volume, domain-adapted language model and a standardized benchmark for measuring legal-domain competence. As a direct consequence, general-purpose Turkish encoders systematically misrepresent specialized legal terminology, fragment domain-specific compound phrases into uninformative subwords, and offer no principled way to quantify how well a model actually understands Turkish legal language.

## 1.1. Motivation

High-resource languages have addressed this gap through domain-adaptive training. LEGAL-BERT [5], for example, was pre-trained on roughly 12 GB of English legal text and consistently beats general-domain baselines on legal tasks. Turkish legal NLP has had no such resource. The shortage has held back Turkish LegalTech: automated tools struggle to parse, retrieve, or reason over the language of Turkish jurisprudence.

This limitation is further compounded by three interrelated challenges specific to applying general-purpose models to Turkish law:

*Lexical and Semantic Divergence (Semantic Shift)*. Legal language leans on specialized jargon and archaic terminology that general models often treat as out-of-vocabulary. Beyond vocabulary gaps, general models drift in *semantic shift*—common words pick up technical meanings inside legal contexts. Given a Turkish inheritance prompt, general models default to “miras” (inheritance) and miss the procedurally correct “tereke” (estate). They likewise confuse “istinaf” (appellate review) with generic terms like “itiraz” (objection) or “karar düzeltme” (correction of decision).

*Tokenization Fragmentation*. Generic tokenizers, tuned for everyday language, mishandle domain compound phrases. On Turkish legal text, they shatter key terminology into uninformative subwords. The appellate phrase “İçtihadı Birleştirme Kararı” breaks into seven fragments under generic Turkish BERT tokenizers; a domain-optimized tokenizer keeps it intact as three meaningful units. Fragmentation does two things at once: it disrupts semantic coherence and inflates sequence length, shrinking the model’s effective context window.

*Syntactic Complexity and Formal Boilerplate*. Supreme Court of Appeals (Yargıtay) decisions and similar texts run long,

*Corresponding author: Mehmet Utku Öztürk, Kalitte Inc., Türkiye. Email: utku@turkhukuk.ai

with deeply nested clauses and recurring boilerplate phrases like "Gereği düşünüldü." Models trained on conversational or journalistic Turkish learn the surface patterns of that grammar, not the reasoning structures legal analysis requires.

### 1.2. Contributions

We address these gaps with HukukBERT, a domain-adaptive legal language model for Turkish law. Our contributions are:

- **Legal Corpus Curation:** We compiled, deduplicated (via MinHash LSH), and balanced a 19 GB Turkish legal training corpus of roughly 2.3 million documents, spanning case law, legislation, and legal scholarship.
- **Domain-Optimized Tokenization:** We trained a 48K WordPiece tokenizer seeded with official legal terminology. It preserves complex legal compound phrases as single units, averaging 4.82 subwords per line versus 6.42 for generic Turkish tokenizers.
- **Hybrid Masking and DAPT:** Our domain-adaptive pre-training (DAPT) pipeline mixes four masking strategies—Whole-Word, Token Span, Word Span, and Keyword Masking over a curated list of 40,000+ legal terms—pushing the model toward deep legal concepts rather than surface morphology.
- **Legal Cloze Test and Top Performance:** We release the Hukuki Cloze Testi (Legal Cloze Test), a 750-question benchmark for Turkish legal-domain adaptation. HukukBERT reaches 84.40% Top-1 accuracy, ahead of the strongest baseline, Mursit-Large, by 5.60 and the widely used BERTurk-Legal encoder by 8.93 absolute percentage points.
- **Downstream Legal Segmentation:** We test HukukBERT on the structural segmentation of official court decisions, formulated as a Beginning-Inside-Outside (BIO) token classification task over government-database sources. HukukBERT hits a 92.8% document pass rate and 99.0% boundary accuracy, ahead of every baseline.

## 2. Related Work

### 2.1. Turkish natural language processing

The trajectory of Turkish NLP has been largely defined by the iterative release of increasingly sophisticated general-domain language models. Early foundational efforts, most notably the BERTurk model suite [6], established strong baselines for Turkish natural language understanding by pre-training standard BERT architectures on diverse, general-purpose corpora, including Wikipedia and filtered web crawls. These models demonstrated that dedicated monolingual pre-training significantly outperforms zero-shot cross-lingual transfer from massively multilingual models.

Recently, the Turkish NLP landscape experienced a significant architectural shift with the introduction of TabiBERT [7]. Built upon the ModernBERT architecture [8], TabiBERT is a state-of-the-art, monolingual Turkish encoder trained from scratch on a rigorously curated 84.88-billion-token corpus. By incorporating contemporary architectural advances—such as Rotary Positional Embeddings (RoPE) and FlashAttention—TabiBERT substantially extends the native context window to 8192 tokens and achieves state-of-the-art results across general Turkish benchmarks (TabiBench).

However, despite these considerable advancements in model capacity and pre-training data volume, general-domain Turkish models exhibit critical performance degradation when subjected to specialized legal texts. Our tokenization analysis reveals that general-purpose architectures inherently struggle to process complex Turkish legal morphology. For instance, generic models such as bert-base-turkish-cased fragment legal texts into an average of 6.42 subwords per line. Remarkably, even the highly capable TabiBERT model suffers the most severe fragmentation in our evaluation, yielding 7.38 average subwords per line on legal corpora.

This excessive fragmentation demonstrates that scaling general-domain data—even to 84 billion tokens—does not inherently resolve the lexical and semantic divergence of Turkish legal terminology. Consequently, when evaluated on domain-specific reasoning benchmarks such as the Legal Cloze Test introduced in this work and detailed in subsequent sections, modern general-domain models including TabiBERT (68.13% Top-1 accuracy) and BERTurk-cased (63.73% Top-1 accuracy) severely underperform. This discrepancy highlights a fundamental limitation in current Turkish NLP: general-domain models cannot effectively process legal text, underscoring the need for DAPT over a custom, domain-specific vocabulary.

### 2.2. Domain-specific pre-training in legal NLP

The limitations of general-domain models have driven a significant paradigm shift toward DAPT across highly specialized fields, including biomedicine (e.g., BioBERT [9]) and science (e.g., SciBERT [10]) [11]. In the legal domain, where semantic precision and logical reasoning are paramount, DAPT has proven essential for capturing the nuances of statutory and jurisprudential language.

A foundational milestone in this area is LEGAL-BERT [5], which demonstrated that models pre-trained on large-scale legal corpora consistently outperform their general-domain counterparts on downstream legal tasks. LEGAL-BERT was pre-trained on a diverse 12 GB English corpus encompassing European Union legislation, United Kingdom court cases, and United States Supreme Court decisions. Crucially, the authors showed that building a custom domain-specific vocabulary and pre-training a model from scratch—or continually pre-training a general checkpoint exclusively on legal data—mitigates the tokenization fragmentation and semantic shift inherent in general-domain models.

As legal language models proliferated, the need for standardized evaluation methodologies became apparent, leading to the introduction of LexGLUE (Legal General Language Understanding Evaluation) [12], a benchmark comprising multiple sub-tasks such as multi-label document classification, term extraction, and legal named entity recognition (NER). LexGLUE provided the empirical framework necessary to demonstrate that domain-specific encoders possess fundamentally superior semantic representation of legal text compared to general-purpose architectures. In parallel, the release of large open legal corpora such as the 256 GB Pile of Law [13] has enabled domain-adaptive training at substantially larger scale.

While English and other high-resource languages have benefited immensely from models like LEGAL-BERT and benchmarks like LexGLUE, the paradigm has since extended to additional jurisdictions: Lawformer [14] for Chinese legal long documents, InLegalBERT [15] for Indian law, and the SaulLM-7B large language model for legal text [16]. Analogous resources, however, remain conspicuously absent in the Turkish legal ecosystem. Prior attempts at Turkish legal language modeling, such as BERTurk-Legal [17] and the more recent LegalTurk [18], were constrained by relatively small corpora, limiting their capacity to capture complexities of Turkish jurisprudence and procedural law.

HukukBERT addresses this critical gap by scaling the pre-training corpus to 19 GB of strictly legal Turkish text and introducing a highly specialized 48K WordPiece tokenizer. Furthermore, in the absence of a comprehensive Turkish equivalent to LexGLUE, we introduce the Legal Cloze Test—a rigorous, 750-question synthetic benchmark designed to empirically validate domain adaptation. By aligning our methodology with the DAPT paradigms established by LEGAL-BERT, while simultaneously addressing the unique morphological demands of Turkish, HukukBERT achieves state-of-the-art results on the two Turkish legal tasks evaluated in this work—the Legal Cloze Test and structural segmentation of court decisions—outperforming every evaluated Turkish general-purpose and legal-domain baseline.

## 3. Methodology

The foundational quality of any domain-specific language model depends critically on two factors: the scale and cleanliness of its pre-training corpus and the morphological efficiency of its tokenizer. To build HukukBERT, we engineered a rigorous data pipeline to transform raw, unstructured legal texts into a highly optimized, domain-adapted training dataset.

### 3.1. Corpus collection

The construction of a robust foundational model requires a pre-training corpus that accurately reflects the diverse linguistic spectrum of its target domain. To capture the full morphological and semantic variance of Turkish law, we initially aggregated a large raw corpus comprising approximately 11 million documents, totaling 27.02 GB of text. This collection spanned multiple legal sub-domains, including Supreme Court of Appeals (Yargıtay) decisions (16.23 GB), Council of State (Danıştay) decisions (2.65 GB), and Regional Court of Appeals (İstinaf) rulings (1.90 GB), as well as Constitutional Court (AYM) cases, local court decisions, formal legislation, and scholarly legal literature (Table 1).

The corpus was collected from:

- Mevzuat Bilgi Sistemi (https://www.mevzuat.gov.tr/)—Presented by T.C. Cumhurbaşkanlığı Genel Sekreterliği Hukuk ve Mevzuat Genel Müdürlüğü, hosting current Turkish legislation and repealed regulations. Publicly available.
- UYAP Mevzuat ve İçtihat Programı (https://mevzuat.adalet.gov.tr/)—Presented by T.C. Adalet Bakanlığı Bilgi İşlem Genel Müdürlüğü, hosting Turkish legislation, court decisions and rulings (case law). Publicly available.
- Yargıtay Karar Arama Motoru (https://karararama.yargitay.gov.tr/)—Presented by T.C. Yargıtay Başkanlığı, hosting all past court rulings. Publicly available.
- Türkiye Barolar Birliği Dergisi (https://tbbdergisi.barobirlik.org.tr/)—Bi-monthly magazine published by Türkiye Barolar Birliği, discussing important topics about law. Publicly available.
- T.C. Adalet Bakanlığı Hukuk Sözlüğü (https://sozluk.adalet.gov.tr/)—Presented by T.C. Adalet Bakanlığı Bilgi İşlem Genel Müdürlüğü, hosting a dictionary for legal terminology. Publicly available.
- Academic papers on the legal domain from various sources.

### 3.2. Data preparation: Cleaning and deduplication strategy

Legal documents inherently contain high volumes of formal repetition, procedural templates, and boilerplate text (e.g., standard appellate conclusions such as "Gereği düşünüldü"). If left unaddressed, these redundant sequences cause models to memorize boilerplate rather than learn deep semantic relationships, while simultaneously inflating training costs.

**Deduplication via MinHash LSH**

To systematically eliminate redundancy without discarding semantic value, we implemented a Locality-Sensitive Hashing (LSH) pipeline utilizing the MinHash algorithm, motivated by evidence that aggressive deduplication improves language-model training [19]. We configured the pipeline with 256 permutations (num_perm = 256) and a similarity threshold of 0.90. This deduplication step identified and removed nearly 2 million highly similar or identical Supreme Court of Appeals (Yargıtay) documents, reducing the raw corpus to a cleaned 24.6 GB (Table 2).

**Sub-Domain Balancing**

Despite this extensive deduplication, analysis of the cleaned 24.6 GB corpus revealed a severe sub-domain imbalance. Yargıtay decisions still accounted for an overwhelming 14.48 GB of the dataset, vastly overshadowing statutory and academic texts. Because general language models are sensitive to data distribution, training on this imbalanced corpus would cause the model to overfit to the repetitive syntactic structures of appellate rulings, thereby degrading its performance on analytical texts or structured legislation.

To mitigate this imbalance and ensure holistic domain adaptation, we applied a sub-domain balancing strategy. We downscaled the overrepresented Yargıtay corpus from 14.48 GB to 3.51 GB and upscaled underrepresented but semantically dense categories—particularly legal scholarly articles, academic theses, and core legislation—to enrich the model's exposure to analytical reasoning and definitional knowledge.

This combined deduplication and balancing methodology yielded a final pre-training corpus, consisting of approximately 2.3 million documents and totaling 18.93 GB (Table 3). This proportional distribution ensures that HukukBERT's semantic representations are equally informed by procedural case law, statutory frameworks, and theoretical legal literature.

### 3.3. Domain-specific tokenization (HukukBERT tokenizer)

The morphological complexity of Turkish, a highly agglutinative language, is significantly amplified in legal texts due to the frequent use of compound noun phrases, archaic Ottoman-Turkish derivations, and specialized terminology [20]. When general-purpose tokenizers are applied to this domain, they frequently fail to recognize these complex structures, resulting in severe semantic fragmentation where cohesive legal concepts are reduced into meaningless subword sequences.

**Tokenizer Training Process**

To construct a tokenizer capable of preserving the morphological integrity of Turkish legal text, we trained a WordPiece tokenizer from scratch on our full 19 GB balanced legal corpus. To ensure that domain-specific concepts were not statistically marginalized during subword merging, we explicitly seeded the tokenizer's initialization vocabulary with verified terminology drawn from the Ministry of Justice Legal Dictionary, guaranteeing that essential legal terms are mapped to single discrete tokens. We set the final vocabulary size to 48,000 (48K) tokens—a strategic expansion from standard 32K Turkish vocabularies—to accommodate the dense legal lexicon while avoiding the embedding matrix overhead of larger 128K models.

**Table 1**
**Raw corpus distribution**

| Field | Content | Doc. count | Size (GB) |
|---|---|---|---|
| MEVZUAT | Mevzuat | 16,114 | 0.31 |
| İÇTİHAT | Yerel Hukuk | 526,432 | 4.58 |
| İÇTİHAT | KYB | 161 | <0.01 |
| HUKUK | Sözlük | 2449 | <0.01 |
| İÇTİHAT | İstinaf | 208,901 | 1.90 |
| İÇTİHAT | Yargıtay | 9,584,616 | 16.23 |
| İÇTİHAT | AYM | 21,185 | 0.43 |
| HUKUK DIŞI | Vikipedi | 294,382 | 0.76 |
| HUKUK | Tez | 154 | 0.02 |
| İÇTİHAT | Danıştay | 329,862 | 2.65 |
| HUKUK | Ceza | 644 | 0.03 |
| HUKUK | TBB | 1981 | 0.09 |
| HUKUK | GSU | 381 | 0.02 |
| TOTAL | | 10,987,262 | 27.02 |

**Table 2**
**Deduplicated corpus distribution**

| Field | Content | Doc. count | Size (GB) |
|---|---|---|---|
| MEVZUAT | Mevzuat | 16,067 | 0.31 |
| İÇTİHAT | Yerel Hukuk | 476,049 | 4.16 |
| İÇTİHAT | KYB | 161 | <0.01 |
| HUKUK | Sözlük | 2371 | <0.01 |
| İÇTİHAT | İstinaf | 193,079 | 1.84 |
| İÇTİHAT | Yargıtay | 7,588,993 | 14.48 |
| İÇTİHAT | AYM | 21,013 | 0.43 |
| HUKUK DIŞI | Vikipedi | 275,203 | 0.62 |
| HUKUK | Tez | 152 | 0.02 |
| İÇTİHAT | Danıştay | 320,330 | 2.60 |
| HUKUK | Ceza | 641 | 0.03 |
| HUKUK | TBB | 1981 | 0.09 |
| HUKUK | GSU | 381 | 0.02 |
| TOTAL | | 8,896,421 | 24.60 |

**Comparative Morphological Evaluation**

To empirically validate our WordPiece tokenizer, we conducted a comparative morphological analysis against a range of alternative strategies, including Unigram language models and pipelines utilizing Zemberek morphological preprocessing.[1] The primary evaluation metric was the average number of subwords generated per line of legal text.

The results confirm the superiority of our domain-adapted approach (Table 4). The HukukBERT 48K WordPiece tokenizer achieved a fragmentation rate of exactly 4.82 subwords per line. By comparison, generic Turkish tokenizers produced substantially higher fragmentation: both bert-base-turkish-cased and turkish-large-bert averaged 6.42 subwords per line, while TabiBERT—despite being trained on 84 billion tokens—suffered the highest fragmentation at 7.38 subwords per line.

**Qualitative Analysis**

This statistical advantage translates directly into improved semantic representation. For example, the appellate phrase "İçtihadı Birleştirme Kararı" (Decision on the Unification of Judgments) is fragmented into seven disjointed subword tokens (Table 5) (e.g., İç · tih · adıĠ · Bir · leŞtirmeĠ · Kar · arı) by advanced general models like TabiBERT. HukukBERT's tokenizer, however, cleanly parses this exact phrase into three semantically complete tokens: İçtihadı · Birleştirme · Kararı. By preserving word-level cohesion and preventing artificial sequence length inflation, our tokenizer maximizes the effective context window and provides a stable foundation for the subsequent DAPT phase.

[1] Akın, A. A., & Akın, M. D. (2007). Zemberek, an open source NLP framework for Turkic languages. Open-source toolkit. https://github.com/ahmetaa/zemberek-nlp

**Table 3**
**Balanced corpus distribution**

| Field | Content | Topic | Doc. count | Size (GB) |
|---|---|---|---|---|
| HUKUK | CEZA | MAKALE | 12,780 | 0.59 |
| HUKUK | GSU | MAKALE | 9525 | 0.50 |
| HUKUK | TBB | MAKALE | 19,810 | 0.90 |
| HUKUK | TEZ | MAKALE | 3800 | 0.54 |
| HUKUK DIŞI | WIKI | GENEL | 68,759 | 0.15 |
| İÇTİHAT | AYM | BİREYSEL | 94,572 | 1.90 |
| İÇTİHAT | AYM | NORM | 52,500 | 1.11 |
| İÇTİHAT | DANIŞTAY | KARAR | 189,254 | 1.33 |
| İÇTİHAT | İSTİNAF | KARAR | 181,136 | 1.79 |
| İÇTİHAT | KYB | KARAR | 3925 | 0.01 |
| İÇTİHAT | YARGITAY | KARAR | 1,339,649 | 3.51 |
| İÇTİHAT | YEREL HUKUK | KARAR | 208,432 | 1.94 |
| MEVZUAT | MEVZUAT | CB KARARI | 3450 | 0.07 |
| MEVZUAT | MEVZUAT | CB KARARNAME | 840 | 0.03 |
| MEVZUAT | MEVZUAT | CB YÖNETMELİK | 24,900 | 1.08 |
| MEVZUAT | MEVZUAT | KANUN | 73,400 | 2.58 |
| MEVZUAT | MEVZUAT | GENEL | 945 | 0.02 |
| MEVZUAT | MEVZUAT | KHK | 11,766 | 0.34 |
| MEVZUAT | MEVZUAT | MÜLGA | 1635 | 0.13 |
| MEVZUAT | MEVZUAT | MÜLGA CBK | 345 | <0.01 |
| MEVZUAT | MEVZUAT | MÜLGA KHK | 315 | 0.02 |
| MEVZUAT | MEVZUAT | TEBLİĞLER | 23,235 | 0.30 |
| MEVZUAT | MEVZUAT | TÜZÜK | 900 | 0.03 |
| MEVZUAT | MEVZUAT | YÖNETMELİK | 2280 | 0.06 |
| TOTAL | | | 2,328,153 | 18.93 |

**Table 4**
**Average subwords per line and vocabulary size**

| Tokenizer | Avg. subwords/line | Vocab size |
|---|---|---|
| hukukbert-base-48k-cased | 4.82 | 48,009 |
| hukukbert-base-42k-nozem | 4.84 | 42,020 |
| bert-base-turkish-128k | 5.16 | 128,000 |
| Mursit-Large | 5.26 | 59,008 |
| hukukbert-base-42k-zem | 5.67 | 42,020 |
| BERTurk-Legal | 5.67 | 128,000 |
| hukukbert-42k-unigram | 6.23 | 42,102 |
| bert-base-turkish-cased | 6.42 | 32,000 |
| turkish-large-bert-cased | 6.42 | 32,000 |
| TabiBERT | 7.38 | 50,176 |

**Note:** Bold indicates the lowest fragmentation rate and the tokenizer introduced in this work.

## 3.4. Vocabulary initialization and weight transfer

Because HukukBERT introduces a novel 48K tokenizer optimized for Turkish legal terminology, the base model's weights required careful recalibration. Initializing a model with a fully randomized embedding matrix discards the foundational linguistic knowledge (e.g., basic grammar and common conjunctions) acquired during general pre-training.

To preserve this foundational knowledge while adapting to the new domain, we performed a vocabulary overlap analysis between the original 128K generic vocabulary of the bert-base-turkish-128k-cased model and HukukBERT's new 48K legal vocabulary. This analysis revealed a 76.7% exact match overlap, corresponding to 36,837 shared tokens. The weights for these shared tokens were directly transferred to HukukBERT.

For the remaining 23.3% of the vocabulary—comprising the newly injected, domain-specific legal terms—we applied mean initialization, setting each new token's embedding to the statistical mean of the existing embedding matrix. This hybrid approach preserves the model's general Turkish language capabilities while providing a stable starting point for learning high-density legal vocabulary.

## 3.5. Innovative masking strategies

Standard masked language modeling (MLM) randomly masks individual subword tokens. In highly agglutinative languages like Turkish, this often results in the model trivially predicting grammatical suffixes (e.g., case markers or pluralizations) rather than learning the core semantic roots. To force the

**Table 5**
**Tokenization fragmentation examples across models**

| Phrase | Model | Tokens |
|---|---|---|
| BÖLGE ADLİYE MAHKEMESİ | hukukbert-base-48k-cased | BÖLGE ADLİYE MAHKEMESİ |
| | bert-base-turkish-cased | BÖ ##L ##GE AD ##LI##YE MAH ##K##EM ##ESİ |
| | BERTurk-Legal | bolg ##e adliye mahkemesi |
| | TabiBERT | BAKL GE GA DL AYE GMAH KEM ESA |
| Kanun Hükmünde Kararnamenin | hukukbert-base-48k-cased | Kanun Hükmünde Kararnamenin |
| | bert-base-turkish-cased | Kanun Hükmünde Kararnam ##enin |
| | BERTurk-Legal | kanun huk ##mund ##e kararnamenin |
| | TabiBERT | Kanun ˙G HÃ¼kmÃ¼nde ˙G Kararnamenin |
| ˙Içtihadı Birle¸stirme Kararı | hukukbert-base-48k-cased | Içtihadı Birle¸stirme Kararı |
| | bert-base-turkish-cased | Iç ##tih ##adı Birle¸s ##tir ##me Kararı |
| | BERTurk-Legal | ictihad ##1 birlestir ##me kararı |
| | TabiBERT | °Ã§tihadÄ G Bir leÅLtirme ˘G Kar arÃ |
| Hukuki Nitelendirme | hukukbert-base-48k-cased | Hukuki Nitelendirme |
| | bert-base-turkish-cased | Hukuk ##i Nit ##elendirme |
| | BERTurk-Legal | hukuki nitelendir ##me |
| | TabiBERT | HukukiĠ Nit elendir me |

**Note:** Special characters in the TabiBERT rows are byte-level tokenizer markers (a leading space is encoded as a distinct glyph) and are reproduced verbatim.

**Table 6**
**Legal Cloze Test model comparison (*N* = 750)**

| Model | Top-1 accuracy (95% CI) | Top-3 accuracy (95% CI) |
|---|---|---|
| HukukBERT | 84.40% [81.63–86.82%] | 98.80% [97.74–99.37%] |
| newmindai/Mursit-Large | 78.80% [75.73–81.57%] | 97.73% [96.40–98.58%] |
| KocLab-Bilkent/BERTurk-Legal | 75.47% [72.26–78.41%] | 96.00% [94.35–97.18%] |
| dbmdz/bert-base-turkish-128k-cased | 71.87% [68.54–74.97%] | 95.20% [93.43–96.51%] |
| boun-tabilab/TabiBERT | 68.13% [64.71–71.37%] | 95.33% [93.58–96.63%] |
| dbmdz/bert-base-turkish-cased | 63.73% [60.23–67.10%] | 93.47% [91.47–95.02%] |
| ytu-ce-cosmos/turkish-large-bert-cased | 61.60% [58.07–65.01%] | 91.20% [88.96–93.02%] |

model to internalize complex legal concepts rather than superficial grammatical patterns, we designed a hybrid masking strategy.

Operating with an overall MLM probability of 0.25, our custom pipeline distributes masking across four targeted paradigms:

- **Whole-Word Masking (20%):** All sub-tokens belonging to a single word are masked simultaneously [21]. This prevents the model from relying on partial subword context to guess the remainder of a legal term, thereby teaching holistic word-level concepts.
- **Token Span Masking (20%):** Adopting the methodology established by SpanBERT [22], this approach masks consecutive blocks of tokens. This forces the model to predict entire phrases based on the surrounding boundary tokens, enhancing its understanding of legal clause structures.
- **Word Span Masking (30%):** Span masking is extended to consecutive word blocks, further challenging the model to reconstruct multi-word legal idioms and procedural phrasing.
- **Keyword Masking (30%):** Utilizing a curated list of over 40,000 legal terms, we selectively target and mask domain-critical tokens, ensuring the model allocates significant representational capacity to the precise terminology required for legal reasoning.

## 3.6. Training infrastructure and hyperparameters

The DAPT phase was executed on an NVIDIA H200 SXM high-performance compute environment, completing in 19 h. The model was pre-trained for 2 epochs using a linear learning rate scheduler and the AdamW optimizer.

To ensure training stability over the 19 GB balanced corpus, we set a peak learning rate of 1e-5 and an effective batch size of 960, achieved via a base batch size of 192 combined with 5 gradient accumulation steps. Throughout the training process, key metrics including Train Loss, Evaluation Loss, Learning Rate decay, and Gradient Norms were monitored to ensure steady convergence. Detailed training progression figures are provided in Appendices A through D. The complete model configuration—including the exact parameter count of 122,961,801, the 48,009-token WordPiece vocabulary, the random seed (42), and the full set of architecture and optimization settings—is consolidated in Appendix F to support end-to-end reproducibility.

## 3.7. Evaluation framework

Evaluating foundational language models in specialized domains requires metrics that specifically probe semantic resolution and domain knowledge, rather than general grammatical fluency. Because the Turkish NLP ecosystem currently lacks a standardized, multi-task legal benchmark on par with LexGLUE [12], LegalBench [23], or the multilingual LEXTREME [24], we designed an intrinsic evaluation framework and open-sourced it on the HuggingFace platform.

**Figure 1**
**Legal Cloze Test Top-1 and Top-3 accuracy comparison ($N$ = 750)**

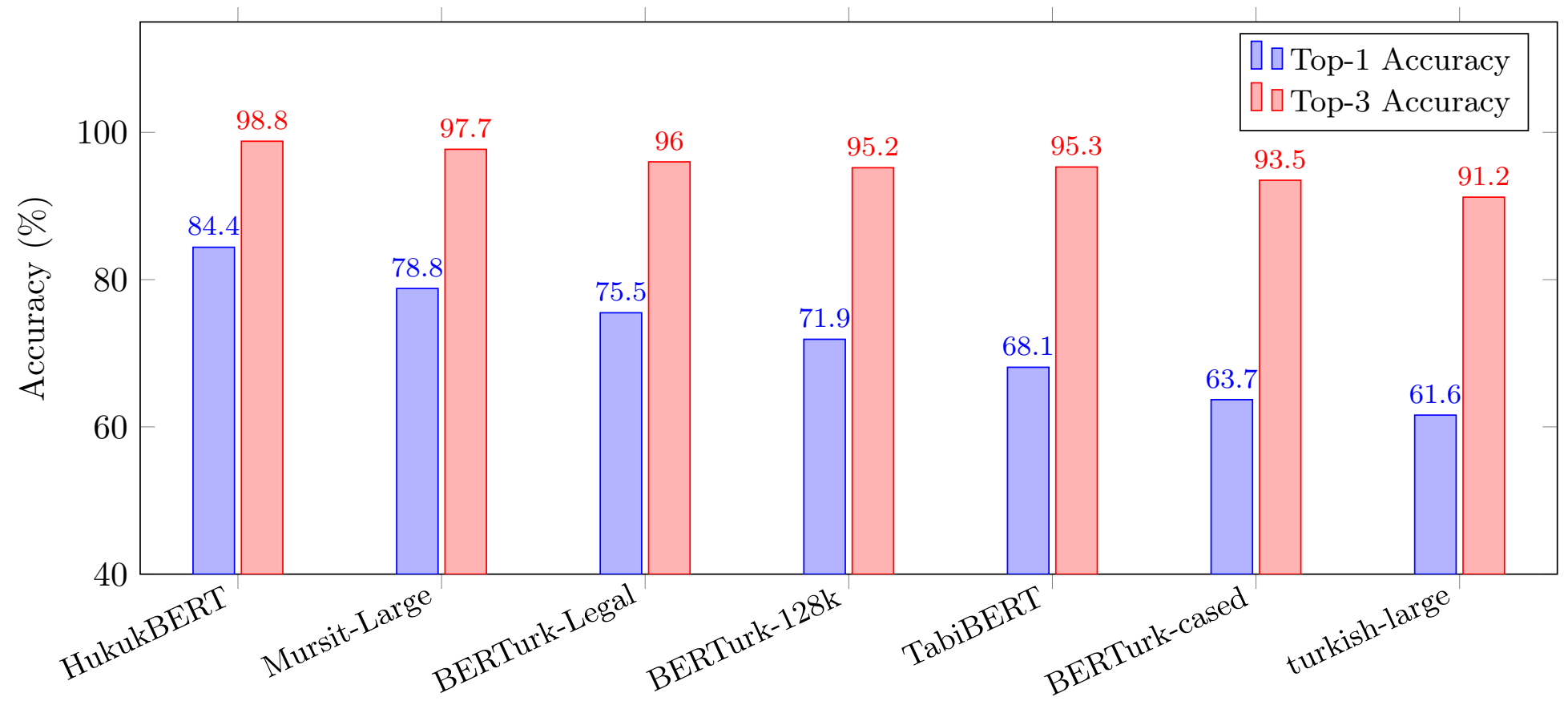

**Intrinsic Evaluation: The Hukuki Cloze Testi (Legal Cloze Test)**

To directly measure the efficacy of our DAPT and hybrid masking strategies, we introduced the Hukuki Cloze Testi (Legal Cloze Test). MLMs are most accurately evaluated intrinsically through their ability to reconstruct masked tokens in highly contextualized environments.

The Legal Cloze Test is a comprehensive, synthetically derived benchmark comprising 750 high-difficulty questions. These questions were sampled across distinct legal sub-domains, including obligations law, criminal law, and procedural law. By requiring models to predict the exact legal term demanded by a specific jurisprudential context, this benchmark serves as a strict probe for legal reasoning and semantic precision.

**Benchmark Generation**. Rather than extracting sentences from existing documents, the 750 questions were synthetically generated using large language models, with each passage strategically constructed to probe a targeted point of Turkish legal knowledge. Every item masks a single legal term whose correct completion is fixed by the surrounding jurisprudential context, and each question is annotated with its law area, mask type, and difficulty level.

**Contamination-Free by Construction**. Because every passage is synthesized specifically for the benchmark rather than drawn from case law, legislation, or scholarship, the test items are absent from the pre-training corpus by construction. This eliminates train–test contamination: no evaluated model, HukukBERT included, could have observed these passages during pre-training.

**Extrinsic Evaluation: Structural Segmentation of Court Decisions**

To demonstrate the practical efficacy of our DAPT, we evaluated HukukBERT on a highly complex downstream LegalTech pipeline: the structural segmentation of official Turkish court decisions. Raw court decisions extracted from government databases are typically monolithic, unstructured text blocks containing interwoven procedural histories, statutory citations, and complex legal reasoning.

The objective is to decompose these unstructured documents into eight discrete, semantically meaningful segments: **header**, **formal**, **claim**, **defense**, **reasoning**, **ruling**, **footer**, and **dissent**. We formulate this as a sequence labeling task utilizing a BIO token classification scheme, requiring the model to identify not only segment spans, but the precise token boundaries where one legal segment transitions into another.

All models were fine-tuned on the v12 segmentation dataset, comprising Turkish court decisions annotated by expert legal professionals. To handle lengthy court decisions, we employed a sliding window approach with a context size of 512 tokens and a stride of 256 tokens. The models evaluated were:

- **HukukBERT** (48K legal vocabulary)
- **BERTurk-128k** (128K general vocabulary)
- **BERTurk-Legal** (128K legal vocabulary)

All models were trained under identical hyperparameters to ensure a fair comparison: a learning rate of 3e-5, an effective batch size of 16, a B-tag weight of 5.0, and automatic class weights optimized over 4 epochs.

## 4. Results

### 4.1. Legal Cloze Test results

We evaluated HukukBERT against a diverse set of baselines, including general-domain Turkish models (TabiBERT, turkish-large-bert, standard BERTurk) and domain-specific Turkish legal models (Mursit-Large and BERTurk-Legal). To ensure statistical robustness and account for variance in prediction confidence, we report both Top-1 and Top-3 accuracies, accompanied by 95% confidence intervals (CI) derived via bootstrap sampling (Table 6). Figure 1 summarises Top-1 and Top-3 accuracy across all evaluated models.

The empirical results demonstrate that HukukBERT achieves state-of-the-art results on Turkish legal court-decision cloze prediction, outperforming all evaluated Turkish general-purpose and legal-domain baselines. HukukBERT achieved a Top-1 accuracy of 84.40% [81.63–86.82%], outperforming the strongest baseline, Mursit-Large (78.80%), by +5.60 absolute percentage points, and the widely used BERTurk-Legal encoder (75.47%) by +8.93.

More critically, the results highlight the limitations of general-domain models. Standard BERTurk-cased achieved only 63.73% Top-1 accuracy, representing a 20.67 percentage-point deficit compared to HukukBERT. Remarkably, the 84-billion-token

TabiBERT model couldn't perform better on this domain-specific task, recording a Top-1 accuracy of 68.13%. This quantitative divide conclusively proves that simply scaling general-domain data cannot substitute for targeted domain adaptation and custom tokenization.

## 4.2. Qualitative error analysis

To understand the mechanics behind HukukBERT's quantitative superiority, we conducted a qualitative error analysis focusing on instances of semantic shift—where common words are incorrectly predicted by general models in place of precise legal terminology—as well as errors in procedural and conceptual legal knowledge.

In the context of inheritance law, models were prompted with spans such as "Mirasbırakanın vefatı ile birlikte..." (Upon the death of the legator...) (Table 7). General models consistently failed to capture the formal legal context, defaulting to colloquial predictions like "miras" (inheritance). In contrast, HukukBERT accurately isolated the highly specific, procedurally correct term "tereke" (estate), assigning it the highest generation probability in its vocabulary and capturing the semantic shift.

Similarly, within strictly procedural contexts, such as execution regimes in criminal law, general models exhibited substantial confusion and relied on generalized grammatical associations. When prompted with a sentence regarding repeat offenders (e.g., "Tekerrür halinde hükmolunan ceza..."), large general-domain models like TabiBERT and even previous legal models like BERTurk-Legal failed to identify the specific statutory procedure, instead predicting generic filler words or incorrect concepts like "genel" or "cezanın." HukukBERT, however, successfully differentiated these regimes, consistently and correctly predicting "mükerrirlere" (to repeat offenders). This qualitative advantage demonstrates that HukukBERT has internalized the strict definitional boundaries and procedural rules of Turkish jurisprudence that typically hinder general-purpose architectures. Table 8 shows the same pattern in commercial law: given "Kural olarak yalnızca [MASK] iflasa tabi kişilerdir," HukukBERT correctly distributes mass over *borçlu* (debtor) and *tacirler* (merchants), while general-domain models default to colloquial terms such as *şirketler* (companies).

## 4.3. Structural segmentation results

The primary evaluation metric is the document pass rate (doc_pass), a strict document-level binary metric where a single incorrect boundary prediction causes the entire document to fail. As detailed in Table 9, HukukBERT achieved a 92.8% doc_pass, outperforming the general-domain 128k model by 8.5 absolute percentage points and BERTurk-Legal by 10.9 absolute percentage points.

A crucial finding during the evaluation process was the divergence between doc_pass and lower-level metrics. For instance, HukukBERT achieved its best doc_pass (92.8%) at epoch 3.67, but its best boundary_f1 (93.0%) slightly earlier at epoch 3.61, and its best collapsed_macro_f1 (95.5%) much earlier at epoch 2.16. This divergence arises because doc_pass requires all boundaries

**Table 7**
**MLM evaluation demonstrating semantic shift resolution in inheritance law**

| Sentence: Mirasbırakanın vefatı ile birlikte, mirasçılara kanun gereği bir bütün olarak geçen malvarlığına [MASK] denir | | |
|---|---|---|
| Model | Prediction (vocab probabilities) | Status |
| turkish-large-bert | miras (0.57) · borç (0.06) · intikal (0.06) | FAIL |
| Mursit-Large | tereke (0.88) · miras (0.06) | WIN |
| BERTurk-Legal | terek (0.12) · mal (0.08) · irat (0.08) | FAIL |
| TabiBERT | miras (0.62) · pay (0.09) · mal (0.08) | FAIL |
| BERTurk-cased | miras (0.83) · servet (0.03) | FAIL |
| BERTurk-128k | miras (0.59) · terek (0.10) · mal (0.07) | FAIL |
| HukukBERT (Ours) | tereke (0.92) · miras (0.04) | WIN |

**Note:** Bold indicates the model introduced in this work.

**Table 8**
**MLM evaluation demonstrating conceptual depth in commercial law**

| Sentence: Kural olarak yalnızca [MASK] iflasa tabi kişilerdir. | | |
|---|---|---|
| Model | Prediction (vocab probabilities) | Status |
| turkish-large-bert | şirketler (0.17) · bankalar (0.09) · işletmeler (0.05) | FAIL |
| Mursit-Large | , (0.06) · şirketler (0.05) · bunlar (0.05) | FAIL |
| BERTurk-Legal | alacaklılar (0.34) · davalılar (0.14) · davacılar (0.03) | FAIL |
| TabiBERT | suç (0.01) · banka (0.01) · iflas (0.01) | FAIL |
| BERTurk-cased | adi (0.20) · bireysel (0.07) · şirketler (0.05) | FAIL |
| BERTurk-128k | hileli (0.06) · bankalar (0.06) · şirketler (0.03) | FAIL |
| HukukBERT (Ours) | borçlu (0.16) · tacirler (0.13) · alacaklılar (0.06) | WIN |

**Note:** Bold indicates the model introduced in this work.

**Table 9**
**Peak downstream performance on the v12 court-decision segmentation dataset**

| Metric | HukukBERT | BERTurk-128k | BERTurk-Legal |
|---|---|---|---|
| Document pass rate (doc_pass) | 92.8% | 84.3% | 81.9% |
| Tolerant document pass (tol_pass) | 96.4% | 88.0% | 89.2% |
| Boundary accuracy (bnd_acc) | 99.0% | 97.2% | 97.6% |
| Boundary F1 (bnd_f1) | 93.0% | 92.4% | 91.9% |
| Span exact F1 (span_exact_f1) | 67.8% | 65.0% | 65.4% |
| Collapsed macro F1 (col_mac_f1) | 95.5% | 93.9% | 94.2% |
| Collapsed weighted F1 | 97.5% | 97.0% | 96.9% |
| Minimum eval loss | 0.1413 | 0.1486 | 0.1441 |

**Note:** Bold indicates the best value for each metric.

within a document to be simultaneously correct. Consequently, the models' overall segmentation quality on a per-token basis may peak earlier and be substantially better than what the strict doc_pass suggests.

## 4.4. Boundary detection and span-level performance

Because sequence boundaries dictate the structural integrity of the parsed document, analyzing the precision and recall of B-tags is crucial (Table 10). Precisely identifying the boundary between the *reasoning* and *ruling* sections is of particular practical importance in court-decision analysis.

**Table 10**
**HukukBERT per-label B-tag precision, recall, and F1 at peak doc_pass (Epoch 3.67)**

| Label | Precision | Recall | F1 score |
|---|---|---|---|
| B-formal | 98.8 | 100.0 | 99.4 |
| B-footer | 99.0 | 99.0 | 99.0 |
| B-defense | 97.4 | 100.0 | 98.7 |
| B-ruling | 94.8 | 99.4 | 97.0 |
| B-dissent | 100.0 | 93.8 | 96.8 |
| B-claim | 92.2 | 95.0 | 93.6 |
| B-reasoning | 89.3 | 96.2 | 92.6 |
| B-header | 52.2 | 100.0 | 68.6 |

The labels can be ranked by difficulty based on HukukBERT's peak performance. The formal, footer, defense, and ruling segments achieve high boundary detection scores (F1 score: 99.4%, 99.0%, 98.7%, and 97.0%, respectively), reflecting their rigid and predictable procedural structure. In contrast, reasoning (F1 score: 92.6%) and claim (F1 score: 93.6%) labels are more challenging due to the length, structural variability, and the semantic diversity that characterize these sections. The header label consistently yields the lowest F1 score (68.6%) across all models. This weakness is because of low precision rather than low recall: while models achieve 100% recall for B-header, they also produce false-positive predictions at the beginnings of unrelated segments.

To evaluate overall span-level classification accuracy beyond boundary detection, Table 11 presents the collapsed (B + I) F1 scores per-segment type.

In the collapsed metrics, the dissent class yields the lowest F1 scores (65–71% range). This is expected, as dissenting opinion sections are rare in the training dataset and exhibit considerable structural variations compared to standard majority rulings.

**Table 11**
**Collapsed (B + I) per-segment F1 at each model's best doc_pass checkpoint**

| Segment | Hukuk-BERT | BERTurk-128k | BERTurk-Legal |
|---|---|---|---|
| Header | 99.3 | 99.4 | 99.3 |
| Formal | 99.8 | 99.4 | 98.7 |
| Claim | 93.9 | 91.0 | 93.2 |
| Defense | 94.4 | 90.6 | 94.5 |
| Reasoning | 97.3 | 97.8 | 97.8 |
| Ruling | 99.6 | 99.4 | 98.9 |
| Footer | 97.9 | 99.8 | 94.7 |
| Dissent | 70.7 | 65.8 | 66.2 |

## 4.5. Convergence, stability, and overfitting dynamics

The convergence analysis reveals important insights into the learning dynamics. HukukBERT demonstrated the fastest convergence profile, indicating that legal-domain pre-training positioned its initial representations in a task-compatible region, enabling high performance with fewer training steps. HukukBERT reached its peak at epoch 3.67, suggesting that 3–4 epochs of training are optimal.

Conversely, BERTurk-Legal exhibited a notable plateau: its best doc_pass (81.9%) was achieved at a very early stage (epoch 0.70). Beyond this point, the model exhibited fluctuating performance that never surpassed this peak. This suggests that while legal pre-training provides a strong initialisation, a large general-purpose vocabulary (128K) that is not aligned to the legal domain does not translate into capacity for complex structural tasks, whereas HukukBERT's smaller but domain-optimised 48K vocabulary does.

Furthermore, all three models exhibited a characteristic overfitting pattern documented in token classification literature: validation loss began to increase early in training while task-level metrics continued to improve. For HukukBERT, minimum cross-entropy loss occurred at epoch 0.64, while the best doc_pass occurred at epoch 3.67 (3.03 epochs apart). This underscores the absolute necessity of selecting production checkpoints based on task-specific metrics rather than validation loss.

## 5. Discussion and Future Directions

HukukBERT shows that Turkish legal NLP cannot be fixed by scaling general-domain data. The empirical pattern is clear: specialized tokenization paired with targeted DAPT is what overcomes semantic shift and morphological fragmentation.

Applying HukukBERT to structural segmentation produced a few key takeaways:

- **Pre-training Quality Tracks Downstream:** HukukBERT's Cloze Top-1 (84.4%) lines up with its segmentation results. Strong domain pre-training pays off on practical tasks.
- **Vocabulary Size Doesn't Beat Domain Alignment:** BERTurk-128k has a 128K vocabulary and still trailed HukukBERT. Raw size is not the issue; how well the vocabulary fits the domain is.
- **Production Readiness:** For Turkish legal segmentation, HukukBERT-base-512-beta is the model we recommend for LegalTech pipelines. Targeted data augmentation for the dissent class is the most useful next step.

Releasing this 19 GB pre-trained encoder gives Turkish NLP a usable foundation. We plan to extend it in three directions—architectural updates, downstream task specialization, and practical deployment.

### 5.1. Architectural evolution: ModernBERT and ColBERT paradigms

Standard BERT gives accurate semantic representations but tops out at 512 tokens. Yargıtay decisions and detailed legislative texts routinely exceed that. The next versions of HukukBERT will move past this limit.

We plan to port our 19 GB balanced corpus and 48K WordPiece tokenizer onto the ModernBERT architecture [8], picking up RoPE and FlashAttention to push the context window to 8192 tokens. We also want to recast HukukBERT's weights into a ColBERT-style late-interaction model. That keeps fine-grained, token-level representations over long documents and supports similarity scoring beyond what sentence-level pooling allows.

### 5.2. Expansion to downstream legal NLP tasks

Pre-training an encoder and evaluating it intrinsically via the Legal Cloze Test is the first phase of domain specialization. The next is fine-tuning HukukBERT on harder downstream tasks.

We plan dedicated models for **Document Segmentation** (parsing boilerplate-heavy court decisions into sections) and **Multi-Label Classification** (tagging documents by overlapping jurisdictions and statutory references). We also want to set baselines for Turkish legal NER—pulling out court names, dates, statutory articles, and monetary values from raw text. We further plan to extend evaluation to legal information retrieval and to conduct systematic ablation studies that isolate the individual contributions of Keyword Masking, the domain-specific tokenizer, and corpus balancing. Benchmarking against recent generative and multilingual legal large language models, together with formal statistical significance testing across tasks, is likewise left to future work; such comparisons will be most informative once a standardized, multi-task Turkish legal benchmark—comparable to LexGLUE for English—becomes available.

### 5.3. Broader impact on Turkish LegalTech research

Computational law in Türkiye has long been held back by a lack of open, high-volume data and models. Releasing HukukBERT—alongside its custom tokenizer and the Legal Cloze Test benchmarking framework—lowers the barrier to entry. Independent researchers, universities, and LegalTech developers can skip the expensive DAPT phase and focus on legal reasoning, document analysis, and language understanding.

## 6. Limitations

HukukBERT is the strongest result we have for Turkish legal language understanding, but it has limitations worth flagging for anyone planning to deploy it.

**Context Window**

Like other BERT-based encoders, HukukBERT caps at 512 tokens. Appellate rulings, multi-page contracts, and full legislative acts routinely run past this. Full-document processing then requires chunking or sliding window techniques, which work fine for local token prediction or sentence-level retrieval but lose the global context of a long legal argument.

**Generative Incapacity (Encoder-Only Design)**

HukukBERT is a bidirectional encoder, built for understanding rather than generation—representation learning, classification, token prediction. It does not produce text. There is no drafting of contracts, no summary generation, no conversational question answering (QA). Its place is upstream, as the semantic backbone or embedding layer for retrieval and extraction pipelines.

**Evaluation Scope**

The Legal Cloze Test probes domain adaptation and semantic shift, but it is the only benchmark we use here. We have not yet quantified HukukBERT's performance on broader real-world tasks like Legal NER or long-document classification. As both the model and the Legal Cloze Test were developed by the same team, we acknowledge the inherent potential for evaluation bias. We note, however, that the synthetic construction of the benchmark—independent of the pre-training corpus—removes the primary mechanism by which such bias typically manifests, namely, train–test contamination.

## 7. Conclusion

Our results show that targeted domain adaptation is crucial for processing Turkish law. By releasing HukukBERT, the 48K WordPiece tokenizer, and the Legal Cloze Test benchmarking framework, we give the Turkish legal NLP community a working starting point. Downstream applications—document classification, NER, semantic retrieval—get easier to build on top, which should accelerate the next generation of Turkish LegalTech.

## Ethical Statement

This study does not contain any studies with human or animal subjects performed by any of the authors.

## Conflicts of Interest

The authors declare that they have no conflicts of interest to this work.

## Data Availability Statement

The Hukuki Cloze Testi (Legal Cloze Test) benchmark is openly available on the Hugging Face Hub at https://huggingface.co/datasets/turkhukuk/hukukbert-cloze-benchmark. The associated evaluation scripts for the benchmark are available at https://github.com/TurkHukuk/hukukbert. The pre-trained HukukBERT model weights and the training code are not publicly released by the authors at this time, though they may be shared privately for research collaboration purposes.

## Author Contribution Statement

**Mehmet Utku Öztürk:** Conceptualization, Software, Data curation, Writing – original draft, Writing – review & editing, Visualization. **Tansu Türkoğlu:** Conceptualization, Software, Data curation, Resources. **Buse Buz-Yalug**: Methodology, Validation, Writing – review & editing, Supervision.

## References

[1] Vaswani, A., Shazeer, N., Parmar, N., Uszkoreit, J., Jones, L., Gomez, A. N., . . . , & Polosukhin, I. (2017). Attention is all you need. In *Advances in Neural Information Processing Systems* (*30*). Curran Associates, Inc.

[2] Devlin, J., Chang, M.-W., Lee, K., & Toutanova, K. (2019). BERT: Pre-training of deep bidirectional transformers for language understanding. In *Proceedings of the 2019 Conference of the North American Chapter of the Association for Computational Linguistics: Human Language Technologies*, 4171–4186. Association for Computational Linguistics.

[3] Liu, Y., Ott, M., Goyal, N., Du, J., Joshi, M., Chen, D., . . . , & Stoyanov, V. (2019). RoBERTa: A robustly optimized BERT pretraining approach. *arXiv Preprint:1907.11692*.

[4] He, P., Liu, X., Gao, J., & Chen, W. (2021). DeBERTa: Decoding-enhanced BERT with disentangled attention. In *International Conference on Learning Representations (ICLR)*.

[5] Chalkidis, I., Fergadiotis, M., Malakasiotis, P., Aletras, N., & Androutsopoulos, I. (2020). LEGAL-BERT: The muppets straight out of law school. In *Findings of the Association for Computational Linguistics: EMNLP 2020* (pp. 2898–2904). Association for Computational Linguistics.

[6] Schweter, S. (2020). BERTurk: BERT models for Turkish (Version 1.0.0) [Software]. *Zenodo*. https://doi.org/10.5281/zenodo.3770924

[7] Türker, M., Kızıloğlu, A., Güngör, O., & Üsküdarlı, S. (2025). TabiBERT: A large-scale ModernBERT foundation model and unified benchmarking framework for Turkish. *arXiv Preprint: 2512.23065*

[8] Warner, B., Chaffin, A., Clavié, B., Weller, O., Hallström, O., Taghadouini, S., . . . , & Poli, I. (2025). Smarter, better, faster, longer: A modern bidirectional encoder for fast, memory efficient, and long context finetuning and inference. In *Proceedings of the 63rd Annual Meeting of the Association for Computational Linguistics*, 2526–2547. https://doi.org/110.18653/v1/2025.acl-long.127

[9] Lee, J., Yoon, W., Kim, S., Kim, D., Kim, S., So, C. H., & Kang, J. (2020). BioBERT: A pre-trained biomedical language representation model for biomedical text mining. *Bioinformatics*, *36*(4), 1234–1240. https://doi.org/10.1093/bioinformatics/btz682

[10] Beltagy, I., Lo, K., & Cohan, A. (2019). SciBERT: A pretrained language model for scientific text. In *Proceedings of the 2019 Conference on Empirical Methods in Natural Language Processing and the 9th International Joint Conference on Natural Language Processing (EMNLP-IJCNLP)*, 3615–3620. Association for Computational Linguistics. https://doi.org/10.18653/v1/D19-1371

[11] Gururangan, S., Marasović, A., Swayamdipta, S., Lo, K., Beltagy, I., Downey, D., & Smith, N. A. (2020). Don't stop pretraining: Adapt language models to domains and tasks. In *Proceedings of the 58th Annual Meeting of the Association for Computational Linguistics*, 8342–8360. Association for Computational Linguistics.

[12] Chalkidis, I., Jana, A., Hartung, D., Bommarito, M., Androutsopoulos, I., Katz, D., & Aletras, N. (2022). LexGLUE: A benchmark dataset for legal language understanding in English. In *Proceedings of the 60th Annual Meeting of the Association for Computational Linguistics (Volume 1: Long Papers)*, 4310–4330. Association for Computational Linguistics.

[13] Henderson, P., Krass, M. S., Zheng, L., Guha, N., Manning, C. D., Jurafsky, D., & Ho, D. E. (2022). Pile of Law: Learning responsible data filtering from the law and a 256GB open-source legal dataset. In *Advances in Neural Information Processing Systems (NeurIPS), Datasets and Benchmarks Track*, *35*.

[14] Xiao, C., Hu, X., Liu, Z., Tu, C., & Sun, M. (2021). Lawformer: A pre-trained language model for Chinese legal long documents. *AI Open*, *2*, 79–84. https://doi.org/10.1016/j.aiopen.2021.06.003

[15] Paul, S., Mandal, A., Goyal, P., & Ghosh, S. (2023). Pretrained language models for the legal domain: A case study on Indian law. In *Proceedings of the 19th International Conference on Artificial Intelligence and Law (ICAIL)*. https://doi.org/10.1145/3594536.3595165

[16] Colombo, P., Pires, T. P., Boudiaf, M., Culver, D., Melo, R., Corro, C., . . . , & Desa, M. (2024). SaulLM-7B: A pioneering large language model for law. *arXiv Preprint:2403.03883*.

[17] Öztürk, C. E., Özçelik, Ş. B., & Koç, A. (2023). A transformer-based prior legal case retrieval method. In *2023 31st Signal Processing and Communications Applications Conference (SIU)*, 1–4. IEEE.

[18] Zeidi, F., Amasyali, M. F., & Erol, Ç. (2024). LegalTurk optimized BERT for multi-label text classification and NER. *arXiv Preprint:2407.00648*.

[19] Lee, K., Ippolito, D., Nystrom, A., Zhang, C., Eck, D., Callison-Burch, C., & Carlini, N. (2022). Deduplicating training data makes language models better. In *Proceedings of the 60th Annual Meeting of the Association for Computational Linguistics (Volume 1: Long Papers)*, 8424–8445. Association for Computational Linguistics.

[20] Toraman, C., Yılmaz, E. H., Şahinuç, F., & Özçelik, O. (2023). Impact of tokenization on language models: An analysis for Turkish. *ACM Transactions on Asian and Low-Resource Language Information Processing*, *22*(4), 1–21. https://doi.org/10.1145/3578707

[21] Cui, Y., Che, W., Liu, T., Qin, B., & Yang, Z. (2021). Pre-training with whole word masking for Chinese BERT. *IEEE/ACM Transactions on Audio, Speech, and Language Processing*, *29*, 3504–3514. https://doi.org/10.1109/TASLP.2021.3124365

[22] Joshi, M., Chen, D., Liu, Y., Weld, D. S., Zettlemoyer, L., & Levy, O. (2020). SpanBERT: Improving pre-training by representing and predicting spans. *Transactions of the Association for Computational Linguistics*, *8*, 64–77.

[23] Guha, N., Nyarko, J., Ho, D. E., Ré, C., Chilton, A., Narayana, A., . . . , & Li, Z. (2023). LegalBench: A collaboratively built benchmark for measuring legal reasoning in large language models. In *Advances in Neural Information Processing Systems (NeurIPS), Datasets and Benchmarks Track*.

[24] Niklaus, J., Matoshi, V., Rani, P., Galassi, A., Stürmer, M., & Chalkidis, I. (2023). LEXTREME: A multi-lingual and multi-task benchmark for the legal domain. In *Findings of the Association for Computational Linguistics: EMNLP 2023*. Association for Computational Linguistics.

# A Appendices

## A. Training Loss

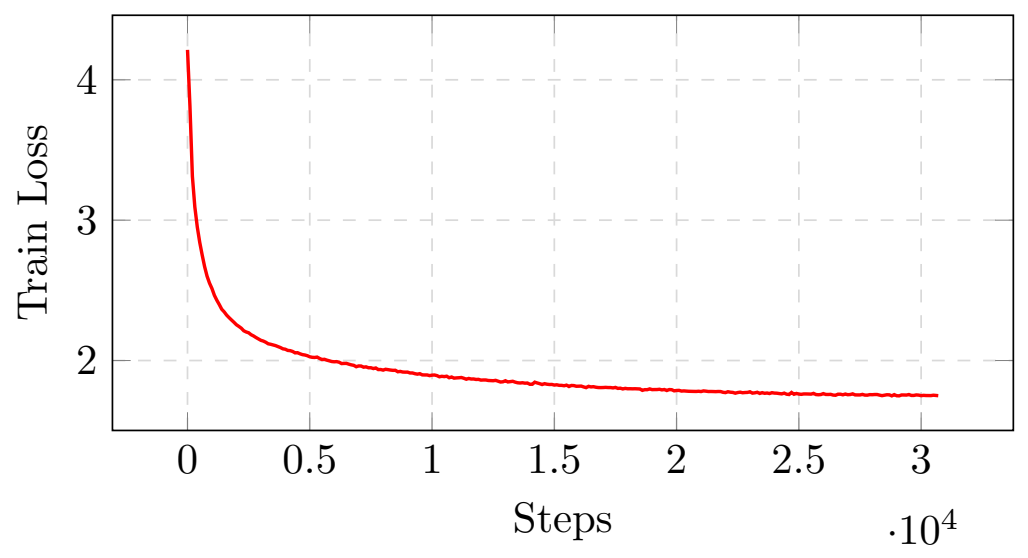

## B. Eval Loss

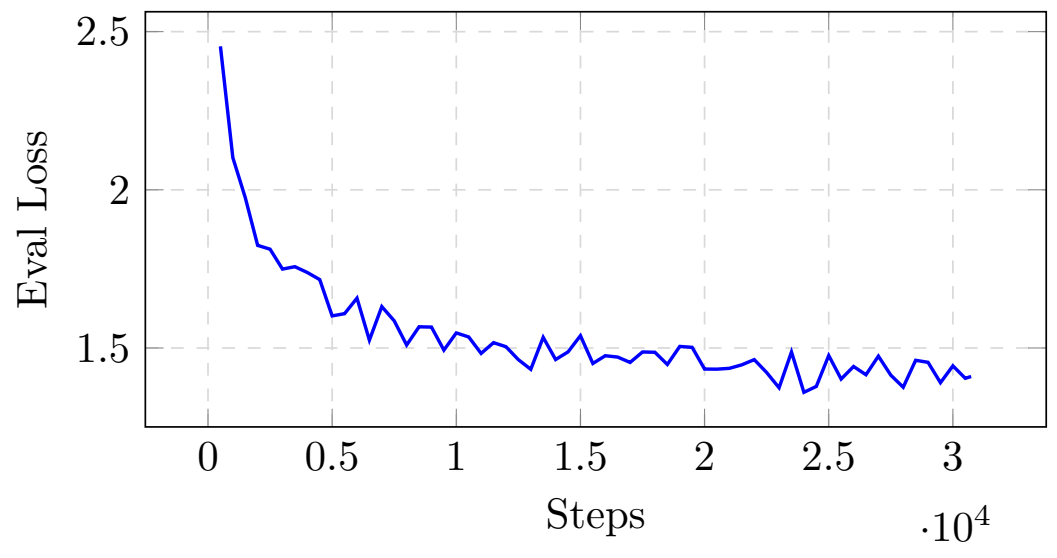

## C. Learning Rate

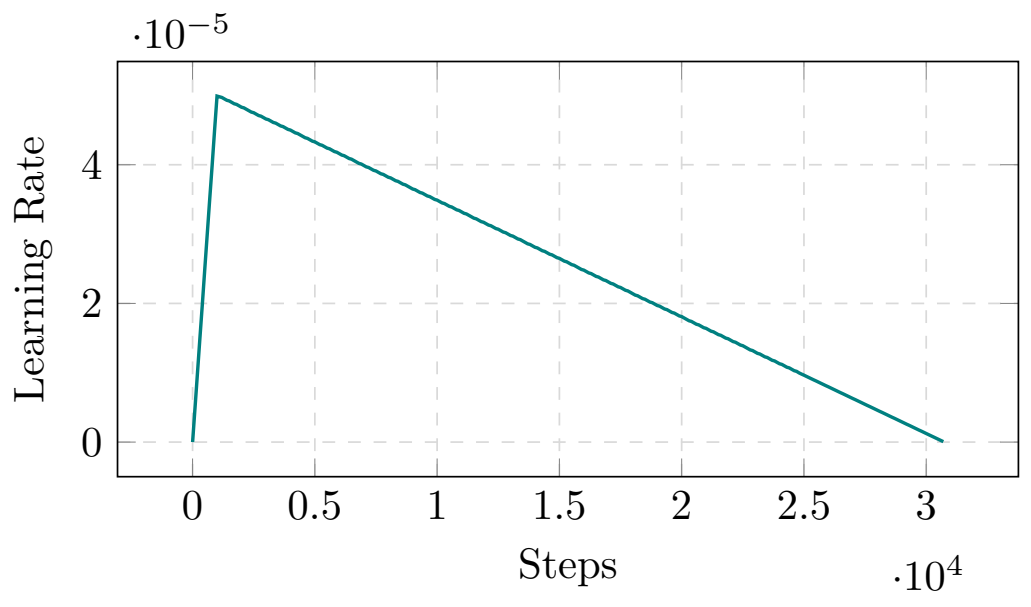

## D. Gradient Norm

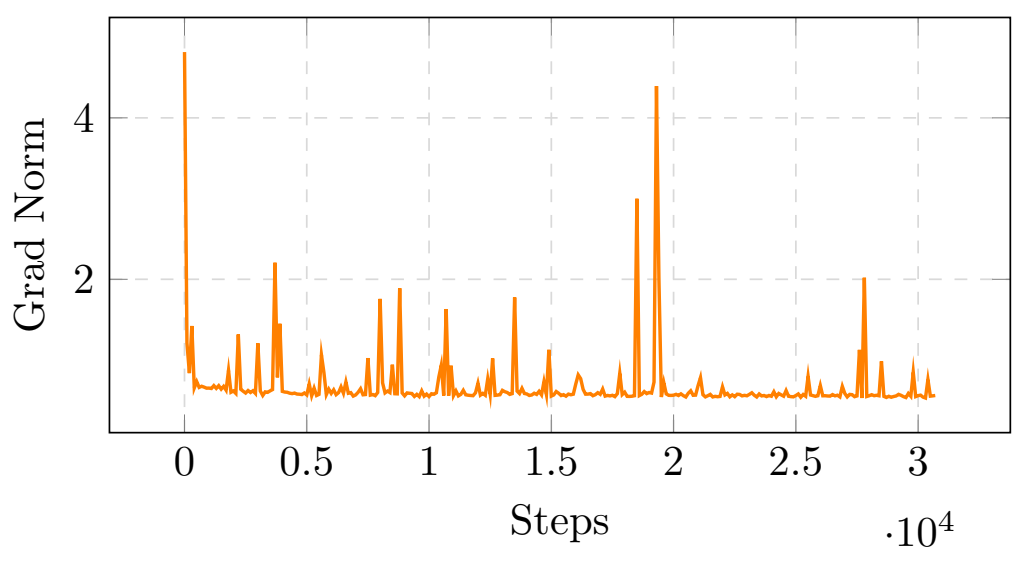

### E. Detailed Epoch-Wise Segmentation Metrics (HukukBERT)

| Epoch | eval_loss | doc_pass | tol_pass | bnd_f1 | span_f1 | col_mac_f1 |
|---|---|---|---|---|---|---|
| 0.25 | 0.1782 | 22.9 | 22.9 | 71.9 | 30.8 | 91.9 |
| 0.50 | 0.1598 | 43.4 | 43.4 | 80.6 | 44.8 | 94.5 |
| 0.75 | 0.1960 | 73.5 | 77.1 | 87.8 | 52.2 | 92.8 |
| 1.00 | 0.2958 | 74.7 | 77.1 | 88.6 | 45.3 | 90.4 |
| 1.25 | 0.2533 | 67.5 | 69.9 | 89.1 | 49.0 | 92.9 |
| 1.50 | 0.1788 | 71.1 | 77.1 | 90.3 | 51.5 | 93.9 |
| 1.75 | 0.1965 | 73.5 | 77.1 | 91.1 | 56.3 | 94.7 |
| 2.00 | 0.2660 | 66.3 | 73.5 | 90.1 | 54.8 | 93.2 |
| 2.25 | 0.1827 | 86.7 | 90.4 | 89.9 | 62.5 | 94.6 |
| 2.50 | 0.2795 | 74.7 | 80.7 | 91.4 | 63.1 | 94.2 |
| 2.75 | 0.2475 | 85.5 | 88.0 | 90.9 | 59.2 | 92.0 |
| 3.00 | 0.1744 | 85.5 | 86.7 | 90.8 | 59.2 | 94.4 |
| 3.25 | 0.2127 | 85.5 | 89.2 | 90.3 | 59.5 | 93.8 |
| 3.50 | 0.3092 | 89.2 | 94.0 | 92.7 | 60.7 | 94.0 |
| 3.67 | 0.3280 | 92.8 | 96.4 | 92.4 | 65.3 | 94.1 |

**Note:** This table reports evaluation checkpoints at 0.25-epoch intervals; the peak values in Table 9 are taken from the full set of logged checkpoints.

### F. Model Configuration and Training Hyperparameters

| Setting | Value |
|---|---|
| *Model architecture* | |
| Backbone | BERT, encoder-only (BertForMaskedLM) |
| Total parameters | 122,961,801 (≈123.0 M) |
| Vocabulary size | 48,009 (WordPiece) |
| Hidden size | 768 |
| Transformer layers | 12 |
| Attention heads | 12 |
| Feed-forward (intermediate) size | 3072 |
| Activation | GELU |
| Hidden/attention dropout | 0.1/0.1 |
| LayerNorm $\varepsilon$ | $1\times10^{-12}$ |
| Max position embeddings | 512 |
| Type vocabulary size | 2 |
| Position embedding type | Absolute |
| Initializer range | 0.02 |
| Tied input/output embeddings | Yes |
| *Domain-adaptive pre-training (DAPT)* | |
| Objective | Masked language modeling (hybrid masking) |
| Optimizer | AdamW |
| Learning-rate schedule | Linear |
| Peak learning rate | $1\times10^{-5}$ |
| Weight decay | 0.01 |
| Epochs | 2 |

(*Continued*)

(*Continued*)

| Setting | Value |
|---|---|
| Effective batch size | 960 (base 192×5 gradient accumulation steps) |
| Maximum sequence length | 512 |
| Random seed | 42 |
| Hardware | 1× NVIDIA H200 SXM |
| Wall-clock time | ≈ 19 h |
| *Hybrid masking (overall MLM probability* = 0.25*)* | |
| Whole-Word Masking | 20% |
| Token Span Masking | 20% |
| Word Span Masking | 30% |
| Keyword Masking | 30% |
| | 40,000+ curated Turkish legal terms |
| *Data* | |
| Training corpus | ~ 2.3 M documents (19 GB, deduplicated and balanced) |
| | Available from the authors on request |
| Legal Cloze Test (intrinsic test set) | 750 items |

*The model configuration, parameter count, and random seed are reported above; the exact train/validation/test split details are available from the authors on reasonable request.*